\documentclass[letterpaper,10pt,conference]{ieeeconf}  
\IEEEoverridecommandlockouts                              
\usepackage{amsmath} 
\usepackage{tikz}
\usetikzlibrary{backgrounds}
\usepackage{adjustbox}

\usepackage{caption}
\usepackage{subcaption}
\usepackage{rotating}
\usepackage{multirow}

\usepackage[innertopmargin=0mm,innerbottommargin=0mm,%
            innerleftmargin= 1.5mm,%
                 leftmargin=-1.5mm,%
           innerrightmargin= 1.5mm,%
                rightmargin=-1.5mm]{mdframed}
\newif\ifmydraft
\mydraftfalse
\definecolor{mydarkred}{RGB}{100,0,0}
{%
\ifmydraft%
\end{mdframed}\vspace{-3mm}%
\par%
\fi%
}

\mydraftfalse

\title{\LARGE \bf
Generation of Human-aware Navigation Maps using \\ Graph Neural Networks}

\author{Daniel Rodriguez-Criado$^{1}$, Pilar Bachiller$^{2}$ and Luis J. Manso$^{1}$
\thanks{$^{1}$Daniel Rodriguez-Criado and Luis J. Manso are with the College of Engineering and Physical Sciences, Aston University, B4 7ET Birmingham, UK
        {\tt\small \{190229717,l.manso\}@aston.ac.uk}}%
\thanks{$^{2}$Pilar Bachiller is with the Robotics and Artificial Vision Laboratory, University of Extremadura, Extremadura, Spain
        {\tt\small pilarb@unex.es}}%
}

\begin{document}

\maketitle
\thispagestyle{empty}
\pagestyle{empty}

\begin{abstract}
Minimising the discomfort caused by robots when navigating in social situations is crucial for them to be accepted.
The paper presents a machine learning-based framework that bootstraps existing one-dimensional datasets to generate a cost map dataset and a model combining Graph Neural Network and Convolutional Neural Network layers to produce cost maps for human-aware navigation in real-time.
The proposed framework is evaluated against the original one-dimensional dataset and in simulated navigation tasks.
The results outperform similar state-of-the-art-methods considering the accuracy on the dataset and the navigation metrics used.
The applications of the proposed framework are not limited to human-aware navigation, it could be applied to other fields where map generation is needed.
\end{abstract}

\section{INTRODUCTION}
Mobile and assistive robots will become widely used in our society as both our companions and assistive aids to older and disabled people with activities of daily living~\cite{Gross2019,Chen2013}.
These robots will be required to follow social conventions to avoid being disruptive, be more predictable and to increase their acceptability amongst people~\cite{KRUSE20131726}.
In order to follow these conventions, robots need to be aware of their surroundings, the people nearby and their activities.
Surveys of the extensive research completed on Human-aware Navigation (HAN) and its impact can be found in~\cite{KRUSE20131726,Rios-Martinez2015} and~\cite{Charalampous2017}.
\par

Human-aware navigation can be seen as a motion planning problem where the robot has to consider many variables.
These variables include the environment, the original robot's pose and its final destination, but also humans, their activity and preferences.
Most approaches use a cost function and a variant of $A^*$~\cite{nilsson2014principles} or Rapidly-exploring Random Trees (RRT)~\cite{lavalle2001rapidly} to search for optimal paths.
Factoring in the power of current computers, the main challenge lies in modelling the aforementioned cost function, one which ideally would consider proxemics, human preferences, activities, emotions and aims.
These functions can be hand-crafted algorithms, Machine Learning (ML) models, or a combination of both.
Hand-crafted algorithms require a considerable amount of resources to develop, are difficult to debug, and even domain experts often overlook meaningful variables.
These solutions often disregard interactions or make simplistic assumptions, whilst ML is able to consider factors that go beyond expert intuition.
The available ML-based approaches also present limitations.
Specifically, they require performing a large numbers of queries when searching for pathways, making the process prohibitively time consuming (see section~\ref{related}).
\par
HAN can also be approached using Reinforcement Learning (RL)~\cite{Chen2017} techniques.
They do not rely on explicit search-based path planning such as $A^*$ or RTT variants, but instead they are end-to-end models that generate the final robot control signals.
Despite having appealing characteristics, end-to-end RL-based HAN algorithms still require reward functions whose development is as challenging as cost functions.
Precisely because RL approaches do not explicitly search, there is no obvious cost function that can be plotted, which means they essentially operate as a black box.
This characteristic makes RL approaches harder to integrate with other algorithms or restrictions including hard-coded conditions (\textit{e.g.}, arbitrary boundaries).
It also makes them more difficult to explain and interpret.
\par
The paper at hand aims to provide a model for robot disruption in human comfortability that can efficiently generate two-dimensional cost maps for HAN considering interactions, an area that has been overlooked until recently.
Although a number of recent works (see section~\ref{related}) address interactions, they perform arbitrary decisions or have serious efficiency limitations when applied to path planning.
\par
The \textbf{contributions} of the paper are two-fold: \textbf{a)} a technique to bootstrap two-dimensional datasets from one-dimensional datasets; and \textbf{b)} \textbf{SNGNN-2D}, an architecture that combines Graph Neural Networks (GNN) and Convolutional Neural Networks (CNN) to generate two dimensional cost maps based on the robot's knowledge.
After training, the resulting ML architecture is able to efficiently generate cost maps that can be used as a cost function for Human-Aware Navigation.
The experiments performed (see section~\ref{sec:experiments}) provide the accuracy of the model, time efficiency and statistical information of the trajectories used by the robot when using SNGNN-2D and a reference Gaussian Mixture Model-based (GMM) algorithm.
The software to bootstrap the two-dimensional dataset and SNGNN-2D has been released as open-source in a public repository, with all the data required to replicate the experiments\footnote{https://github.com/gnns4hri/sngnn2d}.

\section{RELATED WORK}\label{related}
Much of the current literature addressing the problem of robot navigation has focused on the avoidance of collisions with objects, humans and walls, with people being viewed as dynamic obstacles.
Human-aware navigation (HAN) aims to go a step further by considering the underlying social interactions as well as respecting personal and interpersonal spaces. 
In this regard, there have been two major trends in the study of human-aware navigation.
On the one hand, there is proxemics and its psychological and sociological implications~\cite{edward1966hall,Rios-Martinez2015}.
According to~\cite{KRUSE20131726}, there are three main requirements for a robot to navigate socially: comfort, naturalness, and sociability.
Comfort requires respecting interpersonal space, avoidance of collisions and not interfering with peoples' paths.
Naturalness pursuits to mimic human motion and behaviour to make it more predictable.
This requirement is important to reduce the chance of collisions and to lead to more efficient use of space~\cite{KRUSE20131726}.
Finally, sociability implies adherence to social conventions.
For instance, when two people interact, it is expected that no one will interfere with the pathway.
Another example would be to yield to pedestrians in narrow corridors.
Examples of pioneering works that take into account proxemics theory into social navigation are~\cite{Pacchierotti2005} and \cite{kirby2009companion}, where human social conventions, such as personal spaces and human-robot interactions in a hallway are hard-coded as constraints on the robot’s navigation algorithm.
\par

The second trend is Social Force Models (SFM).
The concept was introduced in~\cite{Helbing1995} and continued in~\cite{Truong2017,Patompak2019}. 
SFM and its derivative architectures~\cite{Ferrer2014} model human-robot interactions with repulsive forces, coercing the robot to maintain an interpersonal distance.
These approaches have been used to model the movement of pedestrians so that it can be mimicked by robots.
\par

The main limitation of these trends is that they are pure model-based approaches.
As mentioned in the introduction, the high number of variables to consider in these approaches and the difficulty to estimate their importance make the modelling process especially hard~\cite{VanderHeiden2020}.
\par

The approach presented in this paper and followed by other works covered later in this section is to create models using ML or a combination of hard-coded models and ML.
\par

In ML-based approaches, CNNs have frequently been used to gather information of the whole environment, fed with raw images of the scenario.
An example of this kind of approaches can be found in~\cite{Perez-Higueras2018}, where a Fully Convolutional Network (FCN) learns from expert path demonstrations and is integrated with an RRT* planner.
Their work, although promising, presents some limitations.
The dataset is relatively small due to the tedious task of generating trajectories by humans for each sample.
More importantly, it does not encode significant features for social rules such as human-human and human-object interactions.
\par

Among learning-based models, Deep Reinforcement Learning (DRL) has been used effectively for navigation in crowds.
A policy that allows a robot to navigate through an environment with many pedestrians while respecting the social norms is presented in~\cite{Chen2017}.
In this case, social conventions are modelled by means of a reward function based on geometric features. 
Hence, some of the previously mentioned limitations in model-based methods can also be applied to this approach.
It is important to highlight the difficulty of generating reward functions that encourage social behaviour.
\par

Conversely, Inverse Reinforcement Learning (IRL)~\cite{ng2000algorithms} allows agents to learn from human's experience.
In contrast to reinforcement learning, IRL learns reward functions from human demonstration.
A recent example is~\cite{Sun2020}, which uses a combination of A* and IRL for social navigation taking just the robot's sensors as input.
IRL has also been used to generate social cost maps for robot navigation in~\cite{Vasquez2014}.
Despite IRL-based models achieve outstanding results in HAN, the relations and interactions among humans and human-object are learned implicitly, which can lead to poor performance when there is a dense population.
To solve this problem, \cite{Chen2020crowd} and \cite{chen2019relational} propose the combination of Graph Neural Networks (GNNs) and reinforcement learning for crowd navigation.
They present an end-to end approach where the output of the network is directly the robot control signals.
Thanks to the use of GNNs, the dynamics of crowds are better represented captured. 
However, as highlighted in research relating to this area, the reward function is not learned, instead it is manually designed to keep interpersonal distance to humans.
Additionally, these works do not take into account the symbolic interactions among people.
Thus, the network indicates how much attention agents pay to another but it is not able to determine if they are effectively interacting.

\par
Using an end-to-end approach by means of RL and IRL allows for directly producing the final signals controlling the robot.
Nevertheless, any information of the environment that may affect the robot control has to be initially considered.
This means that, for a HAN system, not only information about the people and their relations with the environment is important, but also the size of the objects, for instance.
It is important to note however that there is no possibility to include that information at a later stage.
An alternative is to build a cost map that can be used by a planner to generate a minimum cost path.
Cost maps can combine information from different sources and are more easily explainable than final control signals, which, from the perspective of the developer, may be very helpful during the development of the system.
\par

SNGNN-2D, the model presented in this paper, is a ML-based approach that aims to generate cost maps for social aware navigation, overcoming some of the aforementioned limitations related to the ability to incorporate symbolic information, such as interactions and other kinds of relationships between entities of the environment.
Specifically, SNGNN-2D combines GNNs and CNNs to generate a cost map from a graph representing the different elements of a room as well as the relationships between them.
An interesting feature of SNGNN-2D is that the model is trained using a map-like 2D dataset bootstrapped from a single-point 1D model developed in a previous work~\cite{manso2020socnav}.
In such 1D model (SNGNN-1D), for any given graph describing a scenario, the network generates a single scalar estimation of how disruptive the robot is overall for the people in the scenario. 
To generate a cost map from these single scores, it would be necessary to query the model for each of the elements in the cost map. 
The time required for generating the cost map this way makes this solution unsuitable for real-time applications.
SNGNN-2D uses a generated dataset that contains scenario-maps tuples which are computed by sampling the output of SNGNN-1D for every robot position.
The bootstrapped dataset is used to train a neural network that produces a final real-time cost map from a given scenario description.

\section{METHOD}\label{method}
\subsection{2D Dataset generation}
The acquisition of two-dimensional cost or \textit{disruption} maps to create datasets for learning purposes generates a number of challenges.
It also requires a significant commitment in comparison to their scalar value counterparts. 
A factor to consider is that the precision of the answer is dependent on the subjects' capability to represent the situation and their preferences graphically.
Their inclination and motivation to stay engaged in the task is an additional challenge. 
\par
From an ML perspective, when factoring in an approximately equal time commitment and effort when generating answers, providing a single scalar for each scenario would yield answers for a higher number of scenarios.
This would in turn generate a higher variability in the input scenarios that would make the model less prone to overfitting.
\par
\begin{figure}[tb]
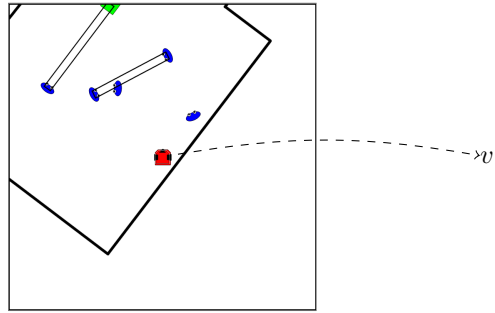
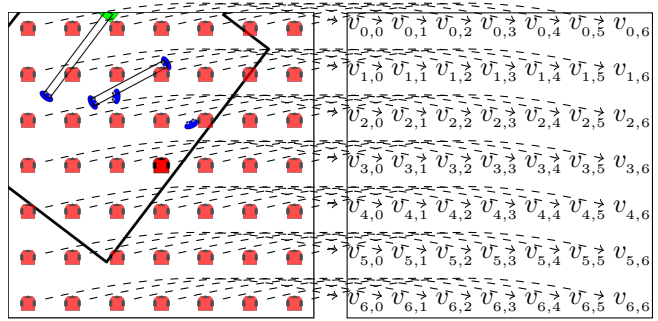

     \begin{subfigure}[t]{\columnwidth}
         \include{tikz/tikz_single_SNGNN1D}
         \vspace*{-4mm}
         \caption{SNGNN-1D can be used to estimate the disruption caused by the robot given a particular scenario.}
         \label{fig:bootstrap-single}
     \end{subfigure}
     \begin{subfigure}[t]{\columnwidth}
         \vspace*{3mm}
         \include{tikz/tikz_grid_SNGNN1D}
         \vspace*{-4mm}
         \caption{The expected 2D outputs are generated performing multiple queries to SNGNN-1D, shifting the scenario around the robot.}
         \label{fig:bootstrap-grid}
     \end{subfigure}
     \caption{The process used to bootstrap the two-dimensional dataset: a) how a single SNGNN-1D query works; b) how to generate two-dimensional outputs.}
     \label{fig:dataset}
\end{figure}

A dataset containing scalars as output data cannot directly be used to train a model which provides two dimensional output, so the approach followed in this case is to use a model which provides one-dimensional value estimations (SNGNN-1D~\cite{Manso2019}) and sample its output shifting the robot's position, bootstrapping this way a two-dimensional dataset.
The process of sampling is depicted in Fig.~\ref{fig:dataset}.
For each scenario in the bootstrapped dataset a matrix of $73x73$ samples is generated.
A total of $37131$ scenarios were randomly generated following the same strategy of SocNav1~\cite{manso2020socnav}.
The dataset split for training, development and test is of $31191$, $2970$ and $2970$ scenarios, respectively.

\subsection{Scenario to graph translation}
\begin{figure}[tb]
\centering
\begin{tikzpicture}[
scale=0.52,
every node/.style={scale=0.52},
xxx/.style={thick,dashed},
midk/.style={very thick},
robotnode/.style= {circle, fill=red!20,  draw=red!60,   very thick,minimum size=4.0mm,inner sep=0pt},
peoplenode/.style={circle,fill=blue!20,  draw=blue!60,  very thick,minimum size=4.0mm,inner sep=0pt},
objectnode/.style={circle,fill=green!20, draw=green!60, very thick,minimum size=4.0mm,inner sep=0pt},
wallnode/.style=  {circle,fill=black!20, draw=black!60, very thick,minimum size=4.0mm,inner sep=0pt},
gridnode/.style=  {circle,fill=gray!20,  draw=gray!60,  very thick,minimum size=4.0mm,inner sep=0pt},
squarednode/.style={rectangle,draw=black!40,inner sep=1.pt}
]


  \node[squarednode,minimum width=6.7cm, minimum height=6.cm,line width=0.2mm](scenario) at (-0.2,0.5){};

  \node[robotnode](robot) at (0,0){};
  \node[peoplenode](p1) at (-2.56,1.55){};
  \node[peoplenode](p2) at (-1.5,1.4){};
  \node[peoplenode](p3) at (-0.98,1.55){};
  \node[peoplenode](p4) at (0.15,2.29){};
  \node[peoplenode](p5) at (0.69,0.9){};
  \node[objectnode](o1) at (-1.22,3.2){};
  
  \draw[->,very thick](p1)--(o1){};
  \draw[->,very thick](p2)--(p4){};
  \draw[->,very thick](p4)--(p2){};


  \node[wallnode](w1) at (-3.22, 3.2){};
  \node[wallnode](w2) at (1.6,3.2){};
  \node[wallnode](w3) at (2.35,2.6){};
  \node[wallnode](w4) at (1.6,1.53){};
  \node[wallnode](w5) at (0.8,0.46){};
  \node[wallnode](w6) at (0.0,-0.61){};
  \node[wallnode](w7) at (-0.85,-1.69){};
  \node[wallnode](w8) at (-1.7,-1.78){};
  \node[wallnode](w9) at (-2.8,-1.){};

  \foreach \x in {-3,...,3}
  {
    \foreach \y in {-3,...,3} 
    {
      \pgfmathsetmacro{\xid}{int(3+\x)}
      \pgfmathsetmacro{\yid}{int(3+\y)}
      \pgfmathtruncatemacro{\label}{int(\yid+100*\xid)}
      \pgfmathsetmacro{\xpos}{1.16*\x*((\y+10)/10)}
      \pgfmathsetmacro{\ypos}{-0.66*\y}
      \node[gridnode](\label) at (\xpos,\ypos-5.5){}; 
     } 
  }
  \foreach \x in {-3,...,3}
  {
    \foreach \y in {-3,...,2}
    {
      \pgfmathsetmacro{\xid}{int(3+\x)}
      \pgfmathsetmacro{\yid}{int(3+\y)}
      \pgfmathsetmacro{\yidd}{int(4+\y)}
      \pgfmathtruncatemacro{\labelxy }{int(\yid+100*\xid)}
      \pgfmathtruncatemacro{\labelyx }{int(\xid+100*\yid)}
      \pgfmathtruncatemacro{\labelxyi }{int(\yidd+100*\xid)}
      \pgfmathtruncatemacro{\labelyix }{int(\xid+100*\yidd)}
      \draw[<->] (\labelxy)--(\labelxyi);
      \draw[<->] (\labelyx)--(\labelyix);
    }
  }

  \draw[->,xxx](robot) to[out=245,in=100 ] (303);
  \draw[->,xxx](o1)
    to[out=120,in=90 ]([xshift={-4.3cm},yshift={+3cm}]robot)
    to[out=270,in=90]([xshift={-4.3cm},yshift={-2.5cm}]robot)
    to[out=270,in=160](200);
  \draw[->,xxx](p1)
    to[out=120,in=90 ]([xshift={-4cm},yshift={+1cm}]robot)
    to[out=270,in=90]([xshift={-4cm},yshift={-2.5cm}]robot)
    to[out=270,in=160](102);
  \draw[->,xxx](p2)
    to[out=200,in=90 ]([xshift={-4.2cm},yshift={-1cm}]robot)
    to[out=270,in=90]([xshift={-4.2cm},yshift={-2.5cm}]robot)
    to[out=270,in=160](202);
  \draw[->,xxx](p3)
    to[out=230,in=90 ]([xshift={-4.2cm},yshift={-1cm}]robot)
    to[out=270,in=90]([xshift={-4.2cm},yshift={-2.5cm}]robot)
    to[out=270,in=160](202);
  \draw[->,xxx](p4)
    to[out=90,in=270 ]([xshift={0.15cm},yshift={3cm}]robot)
    to[out=90,in=90 ]([xshift={3.9cm},yshift={3cm}]robot)
    to[out=270,in=90]([xshift={3.9cm},yshift={-2.5cm}]robot)
    to[out=270,in=20](301);
  \draw[->,xxx](p5)
    to[out=90,in=270 ]([xshift={0.69cm},yshift={3cm}]robot)
    to[out=90,in=90 ]([xshift={3.8cm},yshift={3cm}]robot)
    to[out=270,in=90]([xshift={3.8cm},yshift={-2.5cm}]robot)
    to[out=270,in=10](402);
  \draw[->,xxx](w1)
    to[out=90,in=90 ]([xshift={-4.1cm},yshift={+3cm}]robot)
    to[out=270,in=90]([xshift={-4.1cm},yshift={-2.5cm}]robot)
    to[out=270,in=160](0);
  \draw[->,xxx](w2)
    to[out=50,in=90 ]([xshift={3.7cm},yshift={3.1cm}]robot)
    to[out=270,in=90]([xshift={3.7cm},yshift={-2.5cm}]robot)
    to[out=270,in=40](500);
  \draw[->,xxx](w3)
    to[out=-20,in=90 ]([xshift={3.6cm},yshift={1.5cm}]robot)
    to[out=270,in=90]([xshift={3.6cm},yshift={-2.5cm}]robot)
    to[out=270,in=30](501);
  \draw[->,xxx](w4) to[out=290,in=50](402);
  \draw[->,xxx](w5) to[out=280,in=85] (403);
  \draw[->,xxx](w6) to[out=290,in=80 ] (304);
  \draw[->,xxx](w7) to[out=260,in=100] (205);
  \draw[->,xxx](w8) to[out=270,in=100] (105);
  \draw[->,xxx](w9) to[out=270,in=160] (4);

  \draw[<->,midk](w2)--(w3);
  \draw[<->,midk](w4)--(w3);
  \draw[<->,midk](w5)--(w6);
  \draw[<->,midk](w5)--(w4);
  \draw[<->,midk](w6)--(w7);
  \draw[<->,midk](w8)--(w7);
  \draw[<->,midk](w8)--(w9);

\end{tikzpicture}
\caption{Graphical representation of what the final graph for the scenario depicted in Fig.~\ref{fig:bootstrap-single} looks like.
The node for the room is shown in red (in the centre), blue nodes represent humans, green nodes represent objects and wall nodes are drawn in dark grey.
Grid nodes (in lighter grey) are connected among them and form a squared mesh area.
The nodes in the room graph connect to the closest node of the grid (see dashed arrows).
All nodes in the graph are bidirectionally connected to the room node but are not drawn to facilitate the visualisation.
Edge types are not displayed to avoid clutter.}
\label{fig:graph_transformation}
\end{figure}
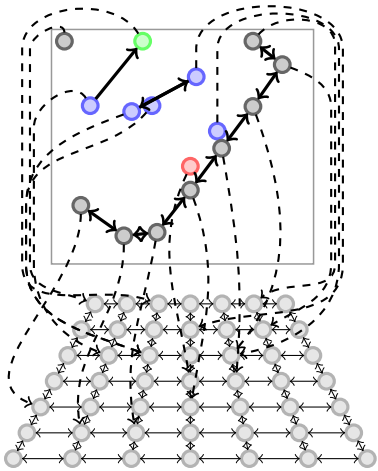

Considering that the input data is not presented in the form of a graph, its conversion to a graph-like structure is one of the most relevant steps if GNNs are to be used.
This process follows the same steps as~\cite{Manso2019}, with the exception that there is an additional grid of $18x18$ nodes whose values are passed to the CNN layers of the architecture and decoded into the final output.
The first part of the graph, which coincides with~\cite{Manso2019} represents the entities in the room and their relations, using a node per entity (room, humans, walls and objects). 
The walls are split into segments, creating a node for each of these segments.
A global \textit{room} node connects to every other node in the graph to facilitate the use of global information in the room using fewer layers.
The human-to-human and human-to-object interactions (if they exist) are represented as edges among the respective nodes.
The first part of the graph can be seen on the top half of Fig.~\ref{fig:graph_transformation}
\par
The grid is a lattice of interconnected nodes, structured to represent the area of the room surrounding the robot by associating them to 2D coordinates.
The number of nodes of this grid and the area they cover are hyperparameters that can be tuned to reach a trade-off between performance, computation time, and area coverage.
The $x$, $y$ coordinates of a grid node in row $i$ and column $j$ from the robot's perspective are computed as:
\begin{align*}
x = \frac{W \; (\left \lfloor{(N-1)/2}\right \rfloor - i)}{N-1} \\
y = \frac{W \; (j - \left \lfloor{(N-1)/2}\right \rfloor)}{N-1}
\end{align*}
where $N$ is the width and height of the lattice and $W$ is the side of the squared area covered by the grid.
\par
Once both sub-graphs have been generated, the elements in the first one are grounded into the grid using additional links to the grid's nodes whose coordinates are the closest (spatial-wise).
This generates a single and final graph that can be fed into the model.
Fig.~\ref{fig:graph_transformation} provides a representation of what the final graph for the scenario depicted in Fig.~\ref{fig:bootstrap-single} looks like.
The GNN layer model used for training allows for labelled edges.
Therefore, each edge is associated with a specific edge type corresponding to the nodes it connects to and their direction.
For instance, a connection from a \textit{human} node to an \textit{object} node will differ that of an \textit{object}-to-\textit{human} connection.
This also applies to the edges between the room graph and the grid graph.
However, the labels of the edges in the grid are labelled based on the direction of the connection to properly account for their relative positions (\textit{i.e.}, up, down, left, right).

Each node on the graph is associated to a feature vector of 21 dimensions.
The input feature vector $h_i^{(0)}$ for the $i$-th node is built by concatenating a one-hot encoding and type-specific metric information: 
$$h_i^{(0)} = (t_i | p_i | o_i | r_i | w_i | g_i)$$
where $t_i$ is a one-hot encoding to differentiate the 5 types of nodes: human, object, room, wall and grid.
The sub-vectors $p_i$, $o_i$ and $w_i$ store metric information for human, object and wall nodes, respectively. 
These metric sub-vectors have 4 dimensions, corresponding to the 2D coordinates of the entity, and the cosine and sine of its orientation. 
The sub-vectors $r_i$ and $g_i$ refer to metric properties of room and grid nodes, respectively. 
Room feature vectors store the number of humans in the room.
The grid vector stores the 2D coordinates of the corresponding node.
Metric vectors which do not correspond to the type of the node are filled with zeros.
\par

\subsection{Architecture description}
\begin{figure*}[tb]
\centering
\includegraphics[width=0.9\textwidth]{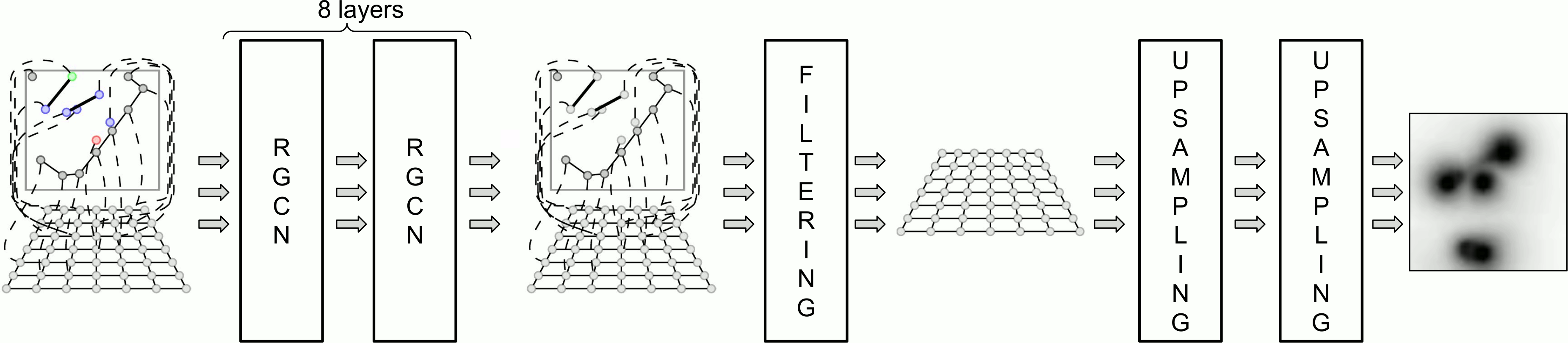}
\caption{Architecture of the SNGNN-2D model.}
\label{fig:architectture}
\end{figure*}

The architecture, depicted in Fig.~\ref{fig:architectture}, has three segments.
The first segment is a sequence of $8$ Relational Graph Convolutional Network layers (R-GCN)~\cite{Schlichtkrull2018}.
Its output is a graph with the same nodes and structure as the input graph, whose feature vectors are converted from $21$ to $7$ dimensions.
The second segment filters out nodes which are not part of the grid and restructures the tensor containing the feature vectors of the grid's nodes into an $18x18x7$ tensor so that it can be used as input to a transposed convolutional layer.
The third and last segment of the architecture is a sequence of two transposed convolutional layers that generate the final $73x73$ output as found in the bootstrapped dataset.
This combination of R-GCN and CNN layers connected through a layer which filters out non-grid nodes to generate cost maps is the second contribution of the paper.

\par

\section{EXPERIMENTS}\label{sec:experiments}
\subsection{Training results}
\begin{figure}[tb]
\centering
  \begin{subfigure}[t]{\columnwidth}
    \includegraphics[width=\textwidth]{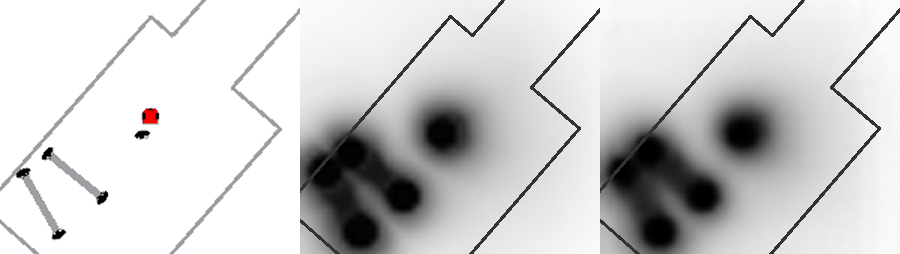}
  \end{subfigure}
  \begin{subfigure}[t]{\columnwidth}
    \includegraphics[width=\textwidth]{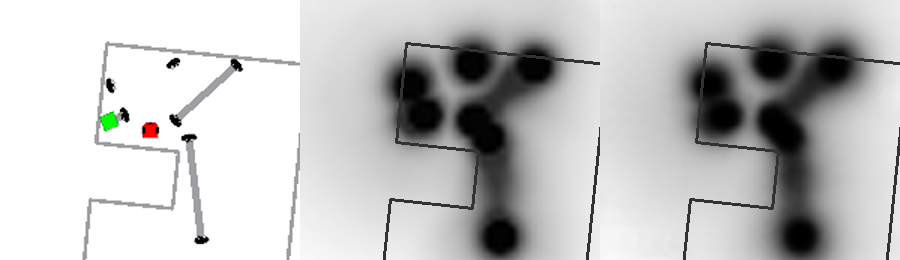}
  \end{subfigure}
  \begin{subfigure}[t]{\columnwidth}
    \includegraphics[width=\textwidth]{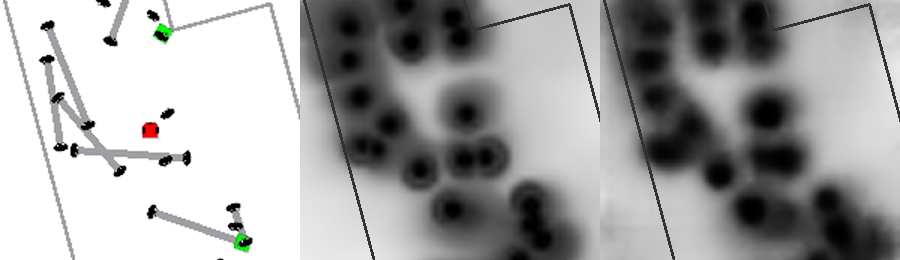}
  \end{subfigure}
  \caption{Results obtained for 3 different scenarios. On the left, representations of the rooms. The central images correspond to the bootstrapped ground-truth. On the right, the output generated by SNGNN-2D.}
  \label{fig:model_results}
\end{figure}

After running $150$ training tasks to optimise the hyperparameters of the architecture, the best model achieved an MSE of $0.00071$, $0.00112$ and $0.00114$ for the training, development and test datasets, respectively.
The model reached the best performance on the development dataset after $35$ epochs.
The $8$ R-GCN layers transform the input feature vectors from $21$ into $7$ dimensions.
The $2$ transposed CNN layers of the model with the lowest test MSE have kernel sizes of $5$ and $3$, with a stride of $2$ and a padding of $1$.
The non-linearity of the best performing model was ELU.
Figure~\ref{fig:model_results} depicts three scenarios and the corresponding output of the SNGNN-1D (sampled) and SNGNN-2D models.

\subsection{Comparison against the reference dataset (SocNav1)}
Given that the bootstrapped dataset does not contain information gathered directly from users but from the output of a one-dimensional GNN model, comparing SNGNN-2D against the test data of the bootstrapped dataset could lead to unrealistic results.
To provide a realistic evaluation of our model with the previous 1D model, the SocNav1/test dataset (one dimensional) was used.
Given that SNGNN-2D provides a whole frame the MSE was computed using only its central pixel, which corresponds to the position of the robot, which is the data that the users assessed on the SocNav1 dataset.
Based on user assessed data only, the MSE computed for the SocNav1/test set was $0.01873$.
The SNGNN-1D version used to generate the bootstrapped dataset performed slightly better than the one reported in the original paper, with an MSE on the SocNav1/test set of $0.0198$.
SNGNN-2D not only achieves better time efficiency, but also achieves a slightly better accuracy on the test set when compared to SNGNN-1D.

\subsection{Real-time evaluation}
The time required by SNGNN-1D to generate a $73x73$ frame in a 6th generation Intel i7 computer with a Geforce GTX 950M is $37.55$ seconds (it requires $5329$ queries to the model).
The time required by SNGNN-2D to generate a similar output is $0.045$ seconds, with just one query.
That is, more than $800$ times faster, three orders of magnitude difference.

\subsection{Comparison against GMM-based methods}
To assess the effectiveness of SNGNN-2D, this section presents simulated navigation results and a comparison with the social aware navigation approach proposed in~\cite{Vega2019}, which is based in Gaussian Mixture Models (GMMs).
\par
The experiments were conducted under simulated environments using SONATA~\cite{baghel2020toolkit}, a toolkit built on top of PyRep~\cite{james2019pyrep} designed to simulate human-populated navigation scenarios and to generate datasets.
SONATA provides an API to generate random scenarios including humans, objects, interactions, the robot and its goals.
The walls delimiting a room are also randomly generated considering rectangular and L-shaped rooms.
\par
SONATA also provides real-time access to the information of the elements in the environment and their properties. This information is used by the two tested methods to generate a cost map, which is integrated in a control system in charge of planning a minimum cost path (using $A*$) and moving the robot towards the goal position.
\par

According to the number of humans in the room, three different types of scenarios were tested: rooms with 2 standing humans and 1 walking human ($S_A$), rooms with 4 standing humans and 2 walking humans ($S_B$) and rooms with 5 standing humans and 3 walking humans ($S_C$).
All the scenarios included a randomly generated number of objects, room shape and wall length. 
The number of interactions between humans or humans and objects was also randomly generated. 
For each type of scenario, each method was executed in 50 different simulations to cover a wide range of situations.
The results of applying each method were evaluated according to the following metrics:
\begin{itemize}
    \item[-] $\tau$: navigation time
    \item[-] $d_t$: travelled distance
    \item[-] $CHC$: cumulative heading changes
    \item[-] $d_{min}$: minimum distance to a human
    \item[-] $si_i$: number of intrusions into the intimate space of humans (closer than $0.45$m)
    \item[-] $si_p$: number of intrusions into the personal space of humans (closer than $1.2$m)
    \item[-] $si_r$: number of intrusions into an interaction (closer than $0.5$m)
\end{itemize}
\par

Tables~\ref{tab:metrics_sngnn2d} and~\ref{tab:metrics_gmm} show the mean and the standard deviation of these metrics using SNGNN-2D and the GMM-based method, respectively, considering separately each group of scenarios. 
For the first two types of scenarios ($S_A$ and $S_B$) results in relation to the mean values of most of the metrics can be considered comparable, although SNGNN-2D produces better results according to the travelled distance ($d_t$) and the cumulative heading changes ($CHC$).
More variability is observed in the GMM-based approach as can be observed by the standard deviation of each parameter.
In addition, for complex scenarios ($S_C$) greater differences can be observed between the two methods, showing that the proposed model behaves in a more socially acceptable way in crowded environments.

\begin{table}[]
\centering
\caption{Navigation results using SNGNN-2D. Mean and Standard Deviation of the metrics for each group of scenarios}
\label{tab:metrics_sngnn2d}
\begin{tabular}{l|ccc}
\hline
Parameter                    & $S_A$                & $S_B$             & $S_C$         \\ \hline
$\tau (s)$                   & 12.11 (4.65)         & 13.75 (5.17)      & 15.99 (6.21)  \\
$d_t (m)$                    & 4.14 (1.85)          & 4.53 (2.20)       & 4.67 (1.91)   \\
$CHC (rads)$                 & 2.87 (1.36)          & 3.18 (1.3)        & 3.64 (1.64)   \\
$d_{min} (m)$                & 1.48 (0.7)           & 1.12 (0.48)       & 1.01 (0.38)   \\
$si_i (\%)$                  & 0 (0)                & 0.11 (0.73)       & 0 (0)         \\
$si_p (\%)$                  & 11.2 (16.5)          & 22.6 (25.77)      & 30.32 (25.4)  \\
$si_r (\%)$                  & 0 (0)                & 0.39 (1.71)       & 0.55 (2.37)   \\
\hline
\end{tabular}
\end{table}

\begin{table}[]
\centering
\caption{Navigation results using the GMM-based method. Mean and Standard Deviation of the metrics for each group of scenarios}
\label{tab:metrics_gmm}
\begin{tabular}{l|ccc}
\hline
Parameter                    & $S_A$            & $S_B$             & $S_C$         \\ \hline
$\tau (s)$                   & 12.16 (7.03)     & 13.12 (6.71)      & 18.94 (10.22) \\
$d_t (m)$                    & 4.92 (2.45)      & 4.88 (2.82)       & 5.7 (3.05)    \\
$CHC (rads)$                 & 4.36 (2.86)      & 4.44 (2.13)       & 5.99 (3.25)   \\
$d_{min} (m)$                & 1.53 (0.67)      & 1.24 (0.51)       & 0.98 (0.39)   \\
$si_i (\%)$                  & 0 (0)            & 0 (0)             & 0.2 (0.91)    \\
$si_p (\%)$                  & 11.98 (27.23)    & 18.06 (27.23)     & 30.16 (32.4)  \\
$si_r (\%)$                  & 0.37 (1.55)      & 0.18 (0.87)       & 1.66 (5.94)   \\
\hline
\end{tabular}
\end{table}

\section{CONCLUSIONS}
As shown in section~\ref{sec:experiments}, the combination of arbitrarily structured and grid nodes in a two staged architecture integrating GNN and CNN layers achieved good results considering both MSE and the properties measured in the navigation simulations.
The reduction in time in comparison to SNGNN-1D is also remarkable ($x800$ speedup), allowing the model to be used in real-time applications.
\par
The main limitation when applying SNGNN-2D is the absence of dynamic properties (\textit{i.e.}, movement) in the input data.
This limitation can be tackled by gathering a dataset which considers the velocity of the robot and humans instead of SocNav1, the one used to bootstrap the 2D dataset.
Another aspect that will be considered for improvement in future works is to train the models to estimate interactions based on previous human behaviour and movements instead of relying on third party models to detect interactions (which are considered given in this work).
\par
Another factor to consider is that the method followed in this paper can be used to generate other kinds of maps with completely unrelated applications by applying the process to other datasets.

\bibliographystyle{IEEEtran}
\bibliography{mybibfile}

\end{document}